\documentclass[letterpaper, 10 pt, conference]{ieeeconf}     

\IEEEoverridecommandlockouts                              

\usepackage{lipsum}  
\usepackage{amsmath,amssymb,amsfonts}
\usepackage{amssymb}
\usepackage{algorithmic}
\usepackage{graphicx}
\usepackage{textcomp}
\usepackage{color}
\usepackage{xcolor}
\usepackage{url}
\usepackage{bm}
\usepackage{hyperref}
\usepackage{acronym}
\usepackage{siunitx}
\usepackage{subcaption}
\hypersetup{
    pdftitle={lol-nmpc}}

\usepackage[textsize=footnotesize]{todonotes}

\newcommand{\Rthree}[0]{\mathbb{R}^3}
\newcommand{\SOthree}[0]{\mathbb{SO}(3)}

\usepackage{tikz}
\usepackage{pgfplots}
\pgfplotsset{compat=1.14}
\usetikzlibrary{backgrounds,arrows,automata,shapes,positioning,calc,through,spy,shapes,shapes.geometric,shapes.multipart,fit,patterns,fadings}
\pgfdeclarelayer{background}
\pgfdeclarelayer{foreground}
\pgfsetlayers{background, main, foreground}

\title{\LARGE \bf
LoL-NMPC: Low-Level Dynamics Integration in Nonlinear Model Predictive Control for Unmanned Aerial Vehicles}

\author{Parakh M. Gupta, Ondřej Procházka, Jan Hřebec, Matej Novosad, Robert Pěnička, Martin Saska
\thanks{The authors are with the Multi-robot Systems Group, Faculty of Electrical
Engineering, Czech Technical University in Prague, Czech Republic (\protect\url{http://mrs.felk.cvut.cz/}). 
This work  has been supported by the Czech Science Foundation (GAČR) under research project No. 23-06162M, by the European Union under the project Robotics and Advanced Industrial Production (reg. no. CZ.02.01.01/00/22\_008/0004590) and by CTU grant no SGS23/177/OHK3/3T/13.
}}

\begin{document}

\newcommand{\position}{p}
\newcommand{\positionvector}{\bm{\position}}
\newcommand{\quaternion}{q}
\newcommand{\orientation}{\bm{\quaternion}}
\newcommand{\linearvel}{v}
\newcommand{\linearvelvector}{\bm{\linearvel}}
\newcommand{\angularvel}{\omega}
\newcommand{\angularvelvector}{\bm{\angularvel}}
\newcommand{\integralerror}{z}
\newcommand{\integralerrorvector}{\bm{\integralerror}}
\newcommand{\statederivative}{\dot{\statestd}}
\newcommand{\statestd}{\bm{x}}
\newcommand{\uavstate}{\statestd}
\newcommand{\statepid}{\bm{x}}
\newcommand{\commandedangularvel}{\angularvel_c}
\newcommand{\propvel}{\Omega}
\newcommand{\propvelvector}{\bm{\propvel}}
\newcommand{\propvelcmd}{\propvel_c}
\newcommand{\propvelcmdvector}{\bm{\propvelcmd}}
\newcommand{\dragforcevector}{\bm{f}_{D}}
\newcommand{\commandedtorquevector}{\bm{\tau_c}}
\newcommand{\torquevector}{\bm{\tau}}
\newcommand{\uavinput}{\bm{u}}
\newcommand{\motorforce}{f}
\newcommand{\thrustvector}{\bm{f}}
\newcommand{\commandedthrustvector}{\bm{f_c}}
\newcommand{\normalizedrpm}{r}
\newcommand{\normalizedrpmvector}{\bm{\normalizedrpm}}
\newcommand{\throttle}{t}
\newcommand{\throttlevec}{\bm{\throttle}}
\newcommand{\commandedcollectivethrottle}{\throttle_c}
\newcommand{\collectivethrust}{T}
\newcommand{\commandedcollectivethrust}{\collectivethrust_c}
\newcommand{\bodythrustvector}{\bm{f}_{T}}
\newcommand{\motorallocationmat}{\bm{M}}
\newcommand{\mixermat}{\bm{G}}
\newcommand{\inertiamat}{\bm{J}}
\newcommand{\mpchorizon}{N}
\newcommand{\errpenmat}{\mathbf{Q}}
\newcommand{\inputpenmat}{\mathbf{R}}
\newcommand{\terminalpenmat}{\mathbf{T}}
\newcommand{\errormat}{\mathbf{\Tilde{\statestd}}}
\newcommand{\errorinputmat}{\mathbf{\Tilde{\uavinput}}}
\newcommand{\statestddesired}{\overset{*}{\statestd}}
\newcommand{\taumot}{k_{mot}}
\newcommand{\thrustcoef}{c_f}
\newcommand{\piderror}{e}
\newcommand{\piderrorvec}{\bm{\piderror}}

\newcommand{\reffig}[1]{Fig.~\ref{#1}}
\newcommand{\refeq}[1]{Eq.~\ref{#1}}
\newcommand{\reftable}[1]{Table~\ref{#1}}
\newcommand{\medialink}{https://mrs.fel.cvut.cz/lol-nmpc}

\acrodef{pid}[PID]{Proportional Integral Derivative}
\acrodef{uav}[UAV]{Unmanned Aerial Vehicle}
\acrodef{uav}[UAV]{Unmanned Aerial Vehicle}
\acrodef{vtol}[VTOL]{Vertical Take-Off and Landing}
\acrodef{mpc}[MPC]{Model Predictive Controller}
\acrodef{nmpc}[NMPC]{Nonlinear Model Predictive Controller}
\acrodef{dfbc}[DFBC]{Differential-Flatness-Based Controller}
\acrodef{indi}[INDI]{Incremental Nonlinear Dynamic Inversion}
\acrodef{ocp}[OCP]{Optimal Control Problem}
\acrodef{imu}[IMU]{Inertial Measurement Unit}
\acrodef{rti}[RTI]{Real-Time Iteration}
\acrodef{rpm}[RPM]{Rotations Per Minute}
\acrodef{rmse}[RMSE]{Root Mean Square Error}
\acrodef{qp}[QP]{Quadratic Programming}
\acrodef{gnss}[GNSS]{Global Navigation Satellite System}

\maketitle
\thispagestyle{empty}
\pagestyle{empty}

\begin{abstract}
	In this paper, we address the problem of tracking high-speed agile trajectories for \acp{uav}, where model inaccuracies can lead to large tracking errors.
	Existing \ac{nmpc} methods typically neglect the dynamics of the low-level flight controllers such as underlying PID controller present in many flight stacks, and this results in suboptimal tracking performance at high speeds and accelerations.
	To this end, we propose a novel \ac{nmpc} formulation, \textit{LoL-NMPC}, which explicitly incorporates low-level controller dynamics and motor dynamics in order to minimize trajectory tracking errors while maintaining computational efficiency.
	By leveraging linear constraints inside low-level dynamics, our approach inherently accounts for actuator constraints without requiring additional reallocation strategies.
    The proposed method is validated in both simulation and real-world experiments, demonstrating improved tracking accuracy and robustness at speeds up to \SI{98.57}{\kilo\meter\per\hour} and accelerations of \SI{3.5}{\g}.
	Our results show an average \SI{21.97}{\percent} reduction in trajectory tracking error over standard \ac{nmpc} formulation, with \textit{LoL-NMPC} maintaining real-time feasibility at \SI{100}{\hertz} on an embedded ARM-based flight computer.
\end{abstract}

\vspace{-0.7em}
\section*{Supplementary Material}
 {\footnotesize
  \vspace{-0.3em}
  \noindent \textbf{Video:} \href{\medialink}{\medialink} 
  \vspace{-0.7em}
 }

\section{Introduction}

\acfp{uav} have garnered considerable research interest due to their remarkable versatility as they can take off and land vertically, loiter at any 3D position, and execute high-speed manoeuvres with extreme agility.
When outfitted with a variety of scanning sensors, UAVs have proven effective in applications such as search-and-rescue~\cite{Rudol_2008_sar}, oil-spill and fire-perimeter monitoring~\cite{Kingston_2008_perimeter_surveillance}, radiation source tracking~\cite{Han2014, Stibinger_2020_radiation_localization}, infrastructure inspection~\cite{Burri_2012_plant_inspection,Merz_2011_inspection}, and aerial coverage scanning~\cite{datsko2024fastCoverage}.
The combination of their versatile capabilities and increasing autonomy is enhancing the efficiency in these operations.

\begin{figure}
	\centering
	\input{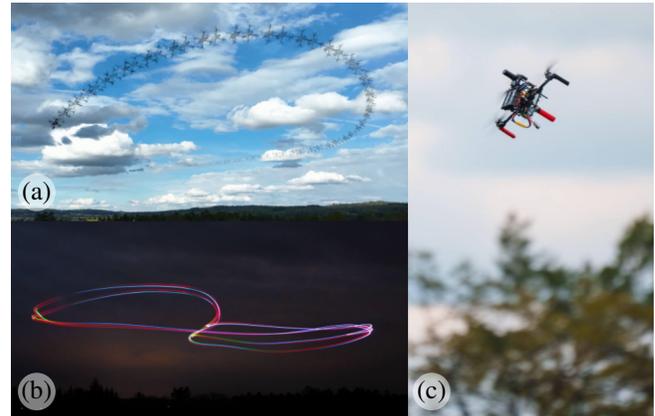}
    \vspace{-1.5em}
	\caption{Illustrative stills from real-world experiments: (a) UAV in flight, (b) long-exposure shot of its trajectory at night, and (c) UAV used in the experiments.\label{fig:intro_picture}}
	\vspace{-1.5em}
\end{figure}

For nearly all of these applications, \acp{uav} are operated at moderate speeds, without pushing them to their performance limits.
In contrast, autonomous drone racing~\cite{Hanover_2024_drone_racing_survey} challenges both the hardware and autonomy capabilities of these vehicles, yet it only serves as a benchmark scenario for applications such as search-and-rescue.
Nevertheless, all the aforementioned applications that use \acp{uav} as autonomous flying sensors can benefit from fast, agile flight, especially with the advent of high-speed cameras and other fast sensors.
However, when \acp{uav} operate at high speeds and execute agile manoeuvres, the risk of significant deviations from planned trajectories (e.g., during inspection missions) increases, potentially jeopardizing the mission by missing the desired position for data collection or by colliding with the environment.
To address this challenge, this paper focuses on onboard control aimed at following a desired trajectory with minimal deviation, that is, to reduce trajectory tracking error, even for agile and fast trajectories (see \reffig{fig:intro_picture}).

Trajectory tracking is a challenging task due to nonlinear \ac{uav} dynamics, propulsion and aerodynamic modeling inaccuracies, actuator dynamics, and delays in state estimation and control loops.
At the same time, the desired trajectories can either consider the feasible but computationally expensive full-scale dynamics model~\cite{foehn_time-optimal_2021}, or a simplified point-mass model which can be quickly computed onboard~\cite{Teissing_2024_PMM_mintime}, but is rather challenging to track due to its partial dynamic infeasibility.
To improve the tracking performance, the \ac{uav} systems~\cite{Foehn22science, Hert2022MRS_system,sun_comparative_2022} often rely on a high-level controller that runs on a companion onboard computer and commands a separate low-level controller which runs on a flight-controller unit.
However, the interaction between the high-level and the low-level controller can cause mismatches, especially in agile flight, and thus complicate precise tracking.
Pushed to their actuator saturation limits, the \acp{uav} can generate larger tracking errors by either lagging behind, overshooting or cutting the corners and due to this cause a collision with any obstacle that the trajectory was planned around.
To fly at the limits of the platform, the high-level controller needs to model the dynamics of the \ac{uav} as close to the real dynamics of the \ac{uav} as possible.

Existing works on high-level control can be broadly categorized into predictive controllers, such as \acf{nmpc}~\cite{nguyen_model_2021}, and non-predictive controllers, including \ac{dfbc}~\cite{Lee2011Geo} and \acf{pid}.
While non-predictive controllers~\cite{Lee2011Geo} are computationally efficient, they rely on feasible trajectories to maintain low tracking errors and often fail to account for the full \ac{uav} dynamics.
In contrast, predictive controllers like \ac{nmpc}~\cite{nguyen_model_2021} leverage modelling of the dynamics to track even partially infeasible trajectories, such as those presented in~\cite{Teissing_2024_PMM_mintime}.
This is also why the research on \ac{nmpc} has primarily focused on improving \ac{uav} dynamics modelling~\cite{nguyen_model_2021}.
Standard NMPC formulations use single-rotor thrust as the control input~\cite{Foehn22science}, while others rely on propeller acceleration~\cite{bicego_nonlinear_2020}.
Additional studies have explored hybrid approaches, integrating \ac{nmpc} with adaptive control~\cite{hanover_performance_2022}, and investigated the impact of low-level and inner-loop controller choices~\cite{sun_comparative_2022}.
However, in many cases, the control inputs required by the low-level controller differs from those modelled within \ac{nmpc}~\cite{foehn_time-optimal_2021,sun_comparative_2022}.
Furthermore, to the best of our knowledge, the true dynamics of low-level controllers have either been entirely neglected in \ac{nmpc} formulations or inaccurately approximated as first-order systems~\cite{nguyen_model_2021}.

To this end, our work introduces a novel \ac{nmpc} which, apart from the standard quadrotor model, also considers both the dynamics of the low-level PID flight controller (that is used in many quadrotor flight stacks such as~\cite{Foehn22science,Hert2022MRS_system}) and the dynamics of the motors themselves.
By leveraging the motor mixing matrix of the low-level flight controller, we accurately model the actuator rise-time dynamics, including PID and motor effects, to enforce both collective and single-rotor constraints directly, and thereby eliminate the need for reallocation strategies for any throttle input.
By modelling the low-level dynamics, the proposed \textit{LoL-NMPC} addresses the model mismatch when using low-level flight controller together with high-level \ac{nmpc} control.


We demonstrate the effectiveness of our approach in enhancing trajectory tracking performance while maintaining computational efficiency in both simulation and real-world experiments.
The \textit{LoL-NMPC} runs at 100 Hz on a small ARM-based flight computer, with only a \SI{56}{\percent} increase in computational time compared to the standard \ac{nmpc} formulation.
Our method has been tested across various trajectory types, consistently delivering improved performance.
The \textit{LoL-NMPC} outperforms standard approaches, achieving an average \SI{21.97}{\percent} reduction in trajectory tracking error while operating outdoors at speeds of up to \SI{98.57}{\kilo\meter\per\hour} and accelerations reaching $3.5$g.

\section{Related Work}


For agile and high-speed flight, choice of the type of controller has been a contested topic due to various advantages and disadvantages of each type of controller.
Among non-predictive controllers, the geometric SE3 controller, proposed in \cite{Lee2011Geo}, is widely used for agile flight throughout the literature and has been very successful in real-world applications.
However, its non-predictive nature prevents it from performing multi-goal optimisation or accounting for actuator saturation.
Predictive controllers were, until recently, constrained by onboard computing, but advances in commercial toolboxes now enable real-time high-speed control~\cite{yutao_chen_matmpc_2019}~\cite{Verschueren2021, Andersson2019}.
Predictive controllers simulate the system to predict the future states using the dynamics model of the \ac{uav}, and coupled with constrained optimisation, they offer the advantage to account for actuator saturation limits and the physical constraints of the platform.
As a result, they handle infeasible trajectories more effectively than non-predictive controllers~\cite{sun_comparative_2022}.
However, to fly at the limits of the platform, the dynamics model inside the \ac{nmpc} needs to be as close to the real dynamics of the \ac{uav} as possible, and including the dynamics and delays originating from the low-level controller becomes crucial to achieve this goal.

Researchers have focused on identifying and modelling the dynamics of the \ac{uav} with increasing detail, for example, by studying aerodynamic effects and propulsion effects.
A seminal work from the authors of \cite{mellinger_minimum_2011} showed that the model of a quadrotor, without the aerodynamics drag, was differentially flat if position and heading were considered as outputs.
Controllers built on this model have been shown to be able to track trajectories with high accuracy and speed.
However, there has been significant research into modelling drag as a first-order effect, and using blade element theory has led to notable performance improvements in both predictive and non-predictive controllers \cite{svacha_improving_2017, mahony_multirotor_2012}.
Authors of \cite{8118153} showed that the dynamics of the \ac{uav} remain differentially flat when drag is modelled as a first-order system.
Identification of such models has been a challenge for traditional methods, and to remedy this, neural-network based approaches have also been presented for identifying and modelling the aerodynamic effects and motor dynamics without increasing the computation time~\cite{bauersfeld_neurobem_2021}.
Another key aspect of the \ac{uav} dynamics is the propulsion system, and the authors of \cite{faessler_thrust_2017} showed that the use of motor modelling can improve the performance of the controller.
Similarly, choice of input has shown to be a key factor in modelling and \cite{bicego_nonlinear_2020} discussed the derivation and use of acceleration of the propellers as an input to the \ac{nmpc} for improved performance on tilt-hex and standard quadrotor \acp{uav}.
Therefore, we tackle this by including the drag characteristics, motor dynamics in our model to improve the tracking performance of the \ac{uav}.

The work presented in \cite{8118153} demonstrated how a \ac{dfbc} can account for drag effects and track trajectories with high accuracy and speed.
For similar agile and high-speed flight, the authors of \cite{sun_comparative_2022} showed that \acp{dfbc} offered 100 times lower compute time in comparison to \ac{nmpc} controllers, and provided better performance than the \ac{nmpc} in case of translational disturbances such as external forces and mass mismatch.
However, in their research, the \ac{nmpc} controllers far outperformed the \ac{dfbc} controllers in trajectory tracking and showed much lower crash rates in cases of rotational disturbances such as centre-of-gravity bias or external moment as well as when tracking infeasible trajectories.
The \ac{dfbc} is shown to primarily suffer due to the inability to account for saturation limits of actuators and physical limits of the platform, as opposed to the \ac{nmpc}.
Work in \cite{sun_comparative_2022} also discussed solving this problem by allocating control using a \ac{qp} solver.
Prior to this, \cite{faessler_thrust_2017} presented that the use of an iterative mixing scheme improves position tracking performance on a \ac{uav} when actuators are saturated or prevented from saturation.
We propose a novel integration of low-level dynamics and linear constraints on the intermediate state which allows us to eliminate the need for reallocation strategies for any throttle input.
Finally, while a significant amount of research has been focused on modelling propulsion and aerodynamic effects, \cite{sun_comparative_2022} highlighted that the choice of inner-loop controller is more crucial than modelling aerodynamic effects, as the performance yields are much higher with \ac{indi}-coupled control schemes.
This was further proven by work in \cite{ezra_tal_accurate_2020}, as the researchers pushed the limits of \ac{dfbc} by combining it with \ac{indi} to achieve high-speed flight of up to \SI{12}{\meter\per\second} at sub-\SI{10}{\centi\meter} accuracy.
Another key area of mismatch was highlighted by \cite{nguyen_model_2021}, where it was discussed that in a cascaded control architecture, inner-attitude controllers are usually approximated as first-order systems although they are second-order systems in reality, and the disturbances accrued from this assumption in modelling require the high-level controller to reject these deviations as disturbances during high-speed flight.
Therefore, we propose a novel method in the next section which bridges the gap between the low-level controller dynamics and the high-level controller dynamics incorporating rate controller dynamics into the \ac{nmpc} formulation.

\section{Problem Formulation}

To understand the modelling problem, we first need to discuss the \ac{uav} dynamics, control input selection, and the standard \ac{uav} model within an \ac{nmpc} framework.


\begin{figure}[h]
	\centering
	\includegraphics[width=0.3\textwidth]{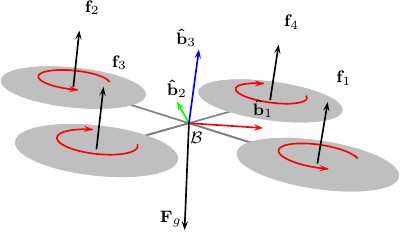}
    \caption{\ac{uav} body frame with motor forces and gravity force.}
    \vspace{-1.0em}
	\label{fig:uav_reference_frame}
\end{figure}

\subsection{Standard \ac{uav} model} \label{sec:standard_model}
The \ac{uav} state is represented by $\statestd=\begin{bmatrix} \positionvector~\orientation~\linearvelvector~\angularvelvector\end{bmatrix}^{T}$ which comprises of position $\positionvector \in \Rthree$, unit quaternion rotation $\orientation \in \SOthree$, velocity $\linearvelvector \in \Rthree$ in world-frame, and body rates of the aircraft (angular velocity of aircraft in body-frame) $\angularvelvector \in \Rthree$.
    The dynamics of the \ac{uav} can be summarized as
\begin{subequations}
	\label{eq:quad_dyn}
	\begin{align}
		\bm{\dot{\position}}         & = \linearvelvector \text{,} \vphantom{\frac{1}{2}} \label{eq:quad_dyn_dot_p}                                                 \\
		\bm{\dot{\quaternion}}       & = \frac{1}{2} \orientation \circledast \begin{bmatrix} 0 \\ \angularvelvector \end{bmatrix}\text{,} \label{eq:quad_dyn_dot_q}                                 \\
        \bm{\dot{\linearvel}}        & = \frac{\bm{R}(\orientation)(\bodythrustvector+\dragforcevector)}{m} + \bm{g}\text{,}\label{eq:quad_dyn_dot_v}               \\
		\bm{\dot{\angularvelvector}} & = \inertiamat^{-1} (\torquevector - \angularvelvector \times \inertiamat \angularvelvector)\text{,}\label{eq:quad_dyn_dot_w}
	\end{align}
\end{subequations}
where the operator $\circledast$ denotes the quaternion multiplication, $\bm{R}(\orientation)$ is the rotational matrix of quaternion $\orientation$, $\bodythrustvector$ is the thrust vector in the body frame, $\dragforcevector$ is the drag force vector in the body frame, $m$ is the mass of the UAV, $\bm{g}$ is the gravitational acceleration, $\inertiamat$ is the diagonal inertial matrix of the rigid body of the \ac{uav}, and $\torquevector$ is the torque produced in the body frame.

Drag force $\dragforcevector$ in~\eqref{eq:quad_dyn} is usually modelled as a linear function of velocity in body frame $\linearvelvector_{\mathcal{B}}$ with drag coefficients $(k_{vx},k_{vy},k_{vz})$~\cite{8118153}, and can be computed as
\begin{equation}
    \normalsize
	\label{eq:drag}
	\dragforcevector = - \begin{bmatrix}k_{vx} v_{\mathcal{B},x} & k_{vy} v_{\mathcal{B},y} & k_{vz} v_{\mathcal{B},z}\end{bmatrix}^{T} \text{.}
\end{equation}
These drag coefficients are identified from the real-world flight by correlating the accelerometer readings with the velocities in body frame.

The derivative of the state \eqref{eq:quad_dyn} can be calculated by first calculating the body thrust vector $\bodythrustvector$ and the torque vector $\torquevector$. 
For this, we can use the motor force vector $\thrustvector$ such that
\begin{equation}
    \normalsize
	\begin{aligned}
		\label{eq:motor_allocation}
		\bodythrustvector & =\begin{bmatrix}0&0&\collectivethrust\end{bmatrix}^T,
            \thrustvector = [f_1, f_2, f_3, f_4]^T, \\
		\begin{bmatrix} \collectivethrust \\ \torquevector \end{bmatrix}
		                  & =
			\begin{bmatrix}
				1           & 1            & 1            & 1          \\
				l/\sqrt{2}  & - l/\sqrt{2} & - l/\sqrt{2} & l/\sqrt{2} \\
				-l/\sqrt{2} & -l/\sqrt{2}  & l/\sqrt{2}   & l/\sqrt{2} \\
				\kappa      & -\kappa      & \kappa       & -\kappa
			\end{bmatrix}
		\thrustvector\text{,}
	\end{aligned}
\end{equation}
where $l$ is the arm length of the symmetric quadrotor frame (\reffig{fig:uav_reference_frame}), and $\kappa$ is the motor torque constant.

\subsection{Standard OCP/MPC formulation}\label{sec:standard_ocp_formulation}

The standard \ac{nmpc} problem is formulated as an \ac{ocp} in discrete time where the goal is to minimize a cost function $J$ over a finite horizon $N$, and therefore, the system dynamics are discretized and propagated using a numerical integrator such as Runge-Kutta 4th order method (RK4). The next state of the system can then be computed as $\statestd_{k+1} = f_{RK4}(\statestd_k,\uavinput_k)$, where $\statestd_k$  and $\uavinput_k$ are the state and the input at time step $k$, respectively, and derivative of the state for each step is computed using the dynamics \eqref{eq:quad_dyn}.


The \ac{nmpc} problem is then formulated as
\begin{equation}
    \normalsize
	\begin{aligned}
		\min_{\uavinput_0,\ldots,\uavinput_{\mpchorizon}} J(&\statestd, \uavinput)  = \sum_{k = 1}^{\mpchorizon-1}\errormat_k^T \errpenmat \errormat_k + \sum_{k = 0}^{\mpchorizon-1}\uavinput_{k}^T \inputpenmat \uavinput_{k} + \errormat_N^T \terminalpenmat \errormat_N \\			\text{subject to :}                                                                                                                                                                                                                                            \\
		\errormat_k                                                               & = \statestddesired_k - \statestd_k,~                                                                                                                                                    
		\statestd_{k+1}                                                            = f_{RK4}(\statestd_k,\uavinput_k),                                                                                                                                                     \\
		\statestd_0                                                               & = \statestd_{initial},~                                                                                                                                                                  
		\uavinput_0                                                                = \uavinput_{initial},                                                                                                                                                                  \\
		\uavinput_k                                                               & \in [\uavinput_{min},\uavinput_{max}],~                                                                                                                                                  
		\angularvelvector_k                                                        \in [\angularvelvector_{min},\angularvelvector_{max}],    
	\end{aligned}
\end{equation}
where $[\bullet]_k$ is the quantity at $t = t_{initial} + k\Delta t$, $\errormat_k$ is the error in the state, $\statestddesired_k$ is the desired state, $\mpchorizon$ is the prediction horizon, $\errpenmat$ and $\inputpenmat$ are the penalty matrices for the state and input respectively, $\terminalpenmat$ is the terminal penalty matrix, $\uavinput_{min}$ and $\uavinput_{max}$ are the control input limits to prevent actuator saturation, and $\angularvelvector_{min}$ and $\angularvelvector_{max}$ represent the limits imposed by state estimators on body rates of the aircraft.
It is also important to note that the computation time of the \ac{ocp} increases with the number of states, inputs, constraints, size of the prediction horizon, as well as the complexity of the dynamics model, and therefore, these quantities act as tuning parameters.

\subsection{Choice of control inputs}\label{sec:choice_of_control_inputs}

As discussed in \cite{bicego_nonlinear_2020}, the true physical limits of the system lie in the maximum torque produced by the motor for any given throttle $\throttle \in [0,1]$, and as the rotational speed of the motor increases, the increase in drag torque prevents further angular acceleration.
It is due to this fact, that the rotational speed of the motor is modelled as a first-order system.
When torque produced by the drag force on the propellers is equal to the maximum torque of the motor, the propeller achieves its maximum rotational speed, and therefore, the maximum force of propulsion.
This maximum force of propulsion along with the maximum torque produced by the motor defines the true physical limits of the propulsion system.

However, the control modality interface for the cascaded control architecture in commercial and research platforms is chosen to be the commanded collective thrust $\commandedcollectivethrust$ and the desired body rates $\angularvelvector_c$ as flight controllers firmwares such as $\text{PX4}$ and $\text{Betaflight}$ expect these as inputs \cite{Foehn22science, Hert2022MRS_system}.
For the \ac{nmpc}, this desired control input can be calculated as part of the first predicted state ($\angularvelvector_c = \angularvelvector_{1} \text{ from } \uavstate_0$) using \eqref{eq:quad_dyn_dot_w} and passed to the low-level controller as an input. Similarly, the desired collective thrust can be calculated as $\commandedcollectivethrust = \sum_{i=1}^{4} (\motorforce_{i,c})$ and passed to the low-level controller as an input.
Owing to this, the control input for the \ac{nmpc} itself is often virtual (non-real in physical sense) and can be chosen in many ways to facilitate modelling fidelity.

\paragraph{First-order motor speed model} When the motor dynamics are chosen to be modelled as a first-order system, the state can be expanded to $\statestd=\begin{bmatrix} \positionvector~\orientation~\linearvelvector~\angularvelvector~\propvelvector\end{bmatrix}^{T}$, and the virtual control input $\uavinput$ can be chosen as the vector of commanded motor speeds $\propvelcmdvector$. The dynamics from \eqref{eq:quad_dyn} can be expanded with the dynamics of the motor as
\begin{equation}
	\begin{aligned}
		\label{eq:motor_dyn_omega}
		\dot{\propvelvector} & = \frac{1}{\taumot}(\propvelcmdvector-\propvelvector) \text{,} \\
		\thrustvector        & = \thrustcoef \propvelvector^{2} \text{,}
	\end{aligned}
\end{equation}
where $\taumot$ is the time constant of the first-order motor dynamics, and $\thrustcoef$ is the thrust coefficient of the propeller. Therefore, $\thrustvector$ can be used to compute the body thrust vector $\bodythrustvector$  and torque vector $\torquevector$ using \eqref{eq:motor_allocation}, and the next state of the system can be obtained according to \eqref{eq:quad_dyn}.

\paragraph{First-order motor force model} When the force produced by each motor is approximated as a first-order system instead, the virtual control input $\uavinput$ can be chosen as the vector of commanded motor forces $\commandedthrustvector$, and the state can be modified as $\statestd=\begin{bmatrix} \positionvector~\orientation~\linearvelvector~\angularvelvector~\thrustvector \end{bmatrix}^{T}$.
In this case, the dynamics of the motor forces can be computed as
\begin{equation}
	\label{eq:motor_dyn_force}
	\dot{\thrustvector} = \frac{1}{k_{mot,f}}(\commandedthrustvector-\thrustvector),
\end{equation}
where $k_{mot,f}$ is the time constant of the first-order motor force dynamics, and \eqref{eq:motor_allocation} can be used to compute the body thrust vector and the torque vector.
However, the force of the actuator is proportional to the square of rotational velocity of the propeller \eqref{eq:motor_dyn_omega}, and the approximation of actuator force to first-order system is inaccurate and causes loss of performance.

\paragraph{No first-order modelling} When first-order effects of the motor dynamics (\eqref{eq:motor_dyn_omega} or \eqref{eq:motor_dyn_force}) are not considered, the virtual control input can be directly chosen as the commanded thrust vector $\commandedthrustvector$, and therefore, \eqref{eq:motor_allocation} can be used to compute the body thrust vector and the torque vector.

\paragraph{Problem of virtual inputs}
In a cascaded control architecture, the low-level controller is usually tasked with tracking the desired body rates and it utilizes the \ac{pid} controller to generate the final commands to the actuators.
Therefore, it is to be noted that none of the aforementioned choices of virtual inputs corresponds to the real input of the low-level controller, and the actual commanded throttles from the low-level controller to the motor controller are obtained through a linear map from the desired body rates to desired throttles as it passes through the \ac{pid} controller.
For example, in our testing, we found that the time taken by the body rates to reach their setpoint (as shown in \reffig{fig:omega_response_comparison}) violates the assumption that the dynamics of the low-level controller are significantly faster than the high-level controller.
Moreover, \cite{faessler_thrust_2017} showed that while choosing the actuator forces or actuator angular velocities as input reduces actuator saturation, conversion to torques and subsequent throttle commands during the mixing process in the low-level controller can cause actuator clipping and re-allocation of control near very-high or very-low throttle values.
In these cases, it becomes imperative to devise a re-allocation and mixing strategy to prevent saturation and gain performance during agile manoeuvres.
In conclusion, it can be stated that the dynamics of the low-level controller play a crucial role in the performance of the \ac{nmpc} controller, and they cannot be neglected without loss of performance.

\section{Proposed LoL-NMPC Methodology}\label{sec:proposed_method}

In this section, we propose a novel method which integrates the dynamics of the low-level controller into the high-level \ac{nmpc} model to significantly improve the tracking performance of the \ac{uav}.
The state of the \ac{uav} is augmented to include the integral error of the \ac{pid} controller and the normalized angular velocities of the motors.
We augment the model by including the \ac{pid} dynamics in state space to formulate a new higher-fidelity model without substantially increasing the computational load of the controller.
The input to the model is chosen as the collective desired throttle and the commanded body rates, which can directly command the low-level controller without causing actuator saturation or performance loss due to propagation of the desired body rates through the \ac{pid} controller or \ac{nmpc} model.
As per the discussion in \ref{sec:choice_of_control_inputs}, rotational velocity of the propeller $\propvel_i$ is a first-order system, and represents the true limit of the motors.
Additionally, for any given propulsion setup, throttle and rotational velocity show a linear relationship such that $f:[0,1]\rightarrow[0,\propvel_{max}]$.
However, we amend this approach and propose an improvement in model fidelity by inculcating the dynamics of the low-level controller inside the model.
First, we extend the state of the \ac{uav} such that
\begin{equation}
	\uavstate^{pid} = \begin{bmatrix} \positionvector~\orientation~\linearvelvector~\angularvelvector~\integralerrorvector~\normalizedrpmvector \end{bmatrix}^{T},
\end{equation}
where $\integralerrorvector$ is the integral error vector for \ac{pid} in $x$, $y$, and $z$ axes, $\normalizedrpmvector$ is the normalized angular velocity vector for each actuator.
It is very important to note that since rotational velocity of the propeller and the applied throttle show a linear relationship for any given input voltage of a motor, normalized angular velocity is equivalent to throttle ($\normalizedrpm \equiv \throttle$), and therefore
\begin{equation}
	\begin{aligned}
		\label{eq:normalized_rpm}
		\normalizedrpm & = \frac{\propvel}{\propvel_{max}},                                  \\
		\thrustvector  & = \motorforce_{max}(\normalizedrpmvector\odot\normalizedrpmvector),
	\end{aligned}
\end{equation}
where $\odot$ denotes element-wise multiplication.
The input to the model $\uavinput_{PID}$ is the collective desired throttle $\commandedcollectivethrottle$ and the commanded body rates $\angularvelvector_c$ such that
\begin{equation}
	\label{input_pid_nmpc}
	\bm{u}^{pid} =  \begin{bmatrix} \commandedcollectivethrottle & \bm{\omega_c} \end{bmatrix}^{T}.
\end{equation}
We then model the PID dynamics in state space (as shown in \cite{Tan2022StateSpacePID}) such that
\begin{equation}
	\begin{aligned}
		\label{eq:pid_equation}
		\piderrorvec               & =  \bm{\omega}_{c} - \bm{\omega},                                            \\
		\dot{\integralerrorvector} & = \piderrorvec ,                                                             \\
		\commandedtorquevector     & = k_{p} \piderrorvec + k_{i} \integralerrorvector,
	\end{aligned}
\end{equation}
where $\commandedtorquevector$ is the commanded torque vector produced by the modelled \ac{pid} controller. 
The $\dot{\piderrorvec}$ is not used in the model because the discrete time-step $\Delta t$ is larger than the required sampling time-step for modelling the transient behaviour of the $\omega$-dynamics using RK4.
For a traditional flight controller, the allocation of individual motor throttle ($\throttlevec$) from torques and collective throttle is performed using a mixer matrix $\mixermat$, such that
\begin{equation}
    \small
	\label{eq:mixer_equation}
	\underbrace{
		\begin{bmatrix}
			\throttle_{c,1} \\
			\throttle_{c,2} \\
			\throttle_{c,3} \\
			\throttle_{c,4} \\
		\end{bmatrix}}_{\throttlevec_c~\text{or}~\normalizedrpmvector_c} = \underbrace{\begin{bmatrix}
			1 & -0.7071 & - 0.7071 & -1.0 \\
			1 & 0.7071  & 0.7071   & -1.0 \\
			1 & -0.7071 & 0.7071   & 1.0  \\
			1 & 0.7071  & - 0.7071 & 1.0  \\
		\end{bmatrix}}_{\mixermat}
	\begin{bmatrix} \commandedcollectivethrottle \\ \commandedtorquevector
	\end{bmatrix}.
\end{equation}

We can then write the motor dynamics as
\begin{equation}
	\label{eq:quad_dyn_thrusts}
	\dot{\normalizedrpmvector} = \frac{1}{k_{mot}}(\normalizedrpmvector_{c}-\normalizedrpmvector), \\
\end{equation}
which we use inside the \textit{LoL-NMPC} quadrotor model instead of the pure motor dynamics~\eqref{eq:motor_dyn_omega} to propagate the state to the next timestep.



\paragraph*{Constraints for \textit{LoL-NMPC}} In addition to low-level dynamics, as mentioned earlier in \ref{sec:choice_of_control_inputs} and \cite{faessler_thrust_2017}, the state-of-the-art suffers from a key issue near throttle limits where actuator clipping and control re-allocation can occur inside the mixer of the low-level controller and cause performance loss during agile manoeuvres.
Our proposed formulation eliminates this key drawback by directly imposing linear constraints in the \ac{ocp} solver in addition to the constraints discussed in \ref{sec:standard_ocp_formulation}.
Therefore, we can impose a linear constraint on the desired normalized angular velocity vector $\normalizedrpmvector_c$ in \eqref{eq:mixer_equation} by expressing it as a function of $\statestd^{pid}$ and $\uavinput^{pid}$, such that
\begin{equation}
	\small
	\begin{aligned}
		\normalizedrpmvector_c & = \mixermat \begin{bmatrix} \commandedcollectivethrottle \\ \commandedtorquevector \end{bmatrix} = \mixermat \begin{bmatrix} \commandedcollectivethrottle \\ k_p(\omega_{cx} - \omega_x) \\ k_p(\omega_{cy} - \omega_y) \\ k_p(\omega_{cz} - \omega_z) \end{bmatrix}, \\
		                       & = \underbrace{\mixermat \begin{bmatrix} 1 & 0      & 0      & 0      \\
				0 & k_{px} & 0      & 0      \\
				0 & 0      & k_{py} & 0      \\
				0 & 0      & 0      & k_{pz}\end{bmatrix}}_{\bm{D}} \uavinput^{pid}  \\
		                       & \underbrace{- \mixermat \begin{bmatrix}
				0 & 0 & \cdots & 0      & 0      & 0      & \cdots & 0 \\
				0 & 0 & \cdots & k_{px} & 0      & 0      & \cdots & 0 \\
				0 & 0 & \cdots & 0      & k_{py} & 0      & \cdots & 0 \\
				0 & 0 & \cdots & 0      & 0      & k_{pz} & \cdots & 0 \\
			\end{bmatrix}}_{\bm{C}} \statestd^{pid},  \\
		\implies               & \normalizedrpmvector_c = \bm{C} \statestd^{pid} + \bm{D} \uavinput^{pid},
	\end{aligned}
\end{equation} and therefore, we can write the constraints on actuators as
\begin{equation}
	\label{eq:normalized_rpm_constraints}
	\begin{aligned}
		\normalizedrpmvector_{min} \leq \bm{C} \statestd^{pid} + \bm{D} \uavinput^{pid} & \leq \normalizedrpmvector_{max}. \\
	\end{aligned}
\end{equation}
This constraint allows us to guarantee torque and thrust allocation in the mixer stage of the low-level controller and improve the performance of the high-level controller at platform limits.

\begin{figure*}[h!]
	\centering
	\begin{tabular}{ccc}
		\includegraphics[width=0.31\textwidth,trim={4.0cm 1cm 6.5cm 1cm},clip]{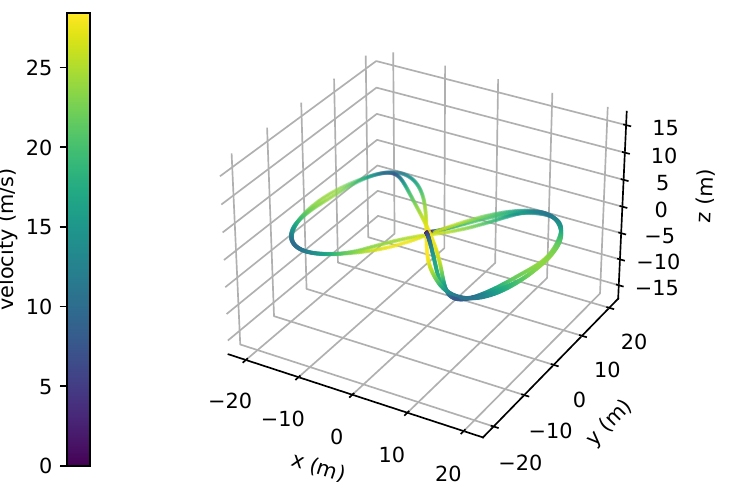}
		 &
		\includegraphics[width=0.31\textwidth,trim={4.0cm 1cm 6.5cm 1cm},clip]{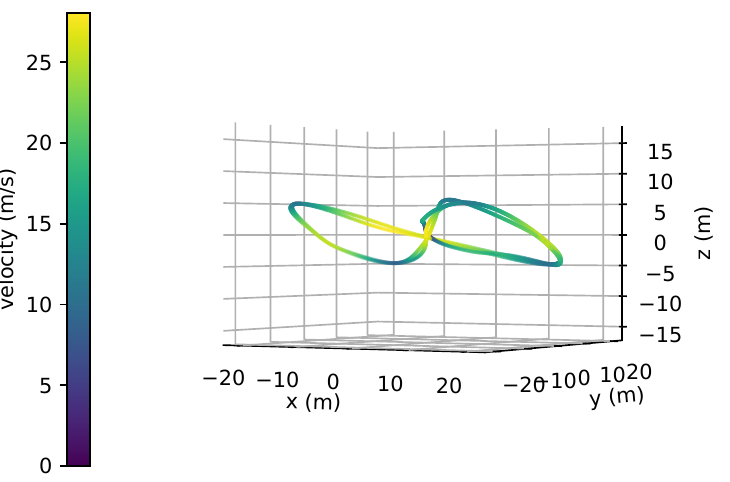}
		 &
		\includegraphics[width=0.31\textwidth,trim={4.0cm 1cm 6.5cm 1cm},clip]{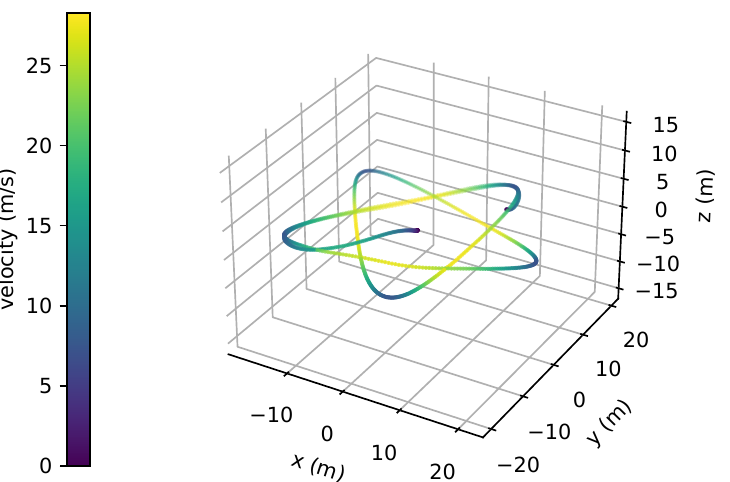}
		\\[-0.3em]
		Figure 8 trajectory (fig 8).
		 &
		Figure 8 slanted trajectory (sl. fig 8).
		 &
		Hypotrochoid trajectory (hyp.).
	\end{tabular}
    \vspace{-0.2em}
	\caption{Our three benchmark trajectories from CPC method~\cite{foehn_time-optimal_2021} which were flown in the real-world.
		\label{fig:real_world_trajectories}}
        \vspace{-1.8em}
\end{figure*}

\section{Experiments and Results}
In this section, we present the methodology and results of our extensive experiments conducted in simulated and real-world flights to validate the performance gained through our novel proposed \textit{LoL-NMPC} formulation.

\subsection{Experimental Setup}

For our simulation and real-world setup, we use a custom-built agile quadrotor measuring \SI{300}{\milli\meter} diagonally (motor-to-motor) and weighing \SI{1.2}{\kilo\gram}.
It is equipped with a CubePilot Cube Orange+ flight controller running \texttt{PX4} firmware which acts as the low-level controller, and is commanded through our open-source MRS system architecture~\cite{Baca_2021} running on a Khadas Vim3 Pro single-board computer.
The \ac{uav} is powered by a 4S Lithium Polymer battery and propelled by a propulsion system capable of producing \SI{80}{\newton} of collective thrust at full throttle, and therefore the \ac{uav} has a thrust-to-weight ratio of $\approx 7$.
For state estimation, the \ac{uav} is equipped with a Holybro F9P RTK GNSS module and receives real-time corrections from a base station, which are fused with the IMU onboard the flight controller to produce odometry for the system.
In the real-world, the motor speed controller sends \ac{rpm} feedback to the flight controller at $\approx \SI{65}{\hertz}$ per motor.
To maintain parity between simulation and real-world experimental setup, the vehicle was tuned for high-agility and aggressive manoeuvres, and its step-response curve for body rates was measured in detail.
The simulation and numerical modelling parameters were then tuned to match the real-world step-response curve of the model predictive controller.
The results of this measurement and tuning are shown in \reffig{fig:omega_response_comparison}, and we would like to highlight through this figure that the settling time of the low-level controller violates the fast dynamics assumption made in the state-of-the-art.

\subsection{Experimental methodology}

\paragraph{Trajectory Generation} It was shown in \cite{sun_comparative_2022}, that non-predictive controllers are sensitive to infeasible trajectories and how predictive controllers outperform them in tracking error on such trajectories.
Therefore, we demonstrate the outstanding performance gains of our approach for three different types of trajectories.
\begin{enumerate}
	\item {CPC: } Authors of \cite{foehn_time-optimal_2021} presented a time-optimal formulation for generation of feasible trajectories by incorporating actuator constraints and real physical parameters of the aircraft.
	      While the method is computationally expensive, it produces near-time-optimal trajectories with high feasibility, and therefore, serves as a good benchmark for high-speed agile flight.
	\item {PMM: } Current state-of-the-art in time-optimal point-mass trajectory generation was presented in~\cite{Teissing_2024_PMM_mintime}. Although point-mass trajectories have low-feasibility for aerial vehicles, it is important to include them for their non-smooth nature that can destabilise the non-linear optimisation for complex models inside the \ac{nmpc} and therefore, they serve as a tough challenge for testing the limits of the platform. 
    While these trajectories can be generated onboard the \ac{uav} in real-time, we use pre-generated trajectories for our experiments.
	\item {Polynomial: } A computationally-efficient method for smooth and feasible polynomial trajectory generation was presented in~\cite{wang2021glst}.
	      These trajectories do not hold the \ac{uav} at its actuator saturation limits but provide a stable benchmark for all types of controllers.
\end{enumerate}

For our experiments, we chose to fly three different types of trajectories (\reffig{fig:real_world_trajectories}), namely, the 8-figure trajectory  (fig 8) for testing horizontal performance, the slanted 8-figure trajectory  (sl. fig 8) for testing both horizontal and vertical performance, and the hypotrochoid trajectory (hyp.) for testing high-speed agile manoeuvres in tight-corners.
Each of these trajectories was generated through the three aforementioned methods for both \SI{2.5}{\g} and \SI{3.5}{\g} accelerations, thereby producing $18$ different trajectory comparisons.

\paragraph{Testing parameters}

From here on, we use the term \textit{'standard'} to refer to the controller implemented using the standard dynamics in \ref{sec:standard_model}, and the standard \ac{ocp} formulation in \ref{sec:standard_ocp_formulation}.
We compare the performance of the standard controller with our proposed controller \textit{LoL-NMPC} which uses the dynamics and constraints described in \ref{sec:proposed_method} inside an \ac{ocp} formulation described in \ref{sec:standard_ocp_formulation}.
We utilise the \ac{rmse} error from the desired trajectory as the primary performance metric for comparison of these controllers, and we also present the computation time statistics for both the simulation and real-world tests.
For simulation tests, we use a direct emulation of our real-world platform and deploy it in Gazebo simulator with \texttt{PX4} running as software-in-the-loop and the \acp{nmpc} running as part of our MRS architecture~\cite{Baca_2021} on \texttt{ROS1}.
For real-world tests, the platform was flown in the open-air in non-windy conditions, and over 180 flights were conducted to ensure error-free implementation of both the standard and our proposed controller.
For both environments, the vehicle was limited to body rates of \SI{6}{\radian\per\second} in each axis and accelerations of \SI{4.0}{\g} to respect state estimator limits for civilian \ac{gnss} use.


\begin{figure}[h!]
	\begin{subfigure}[b]{0.5\textwidth}
		\centering
		\includegraphics[width=\textwidth]{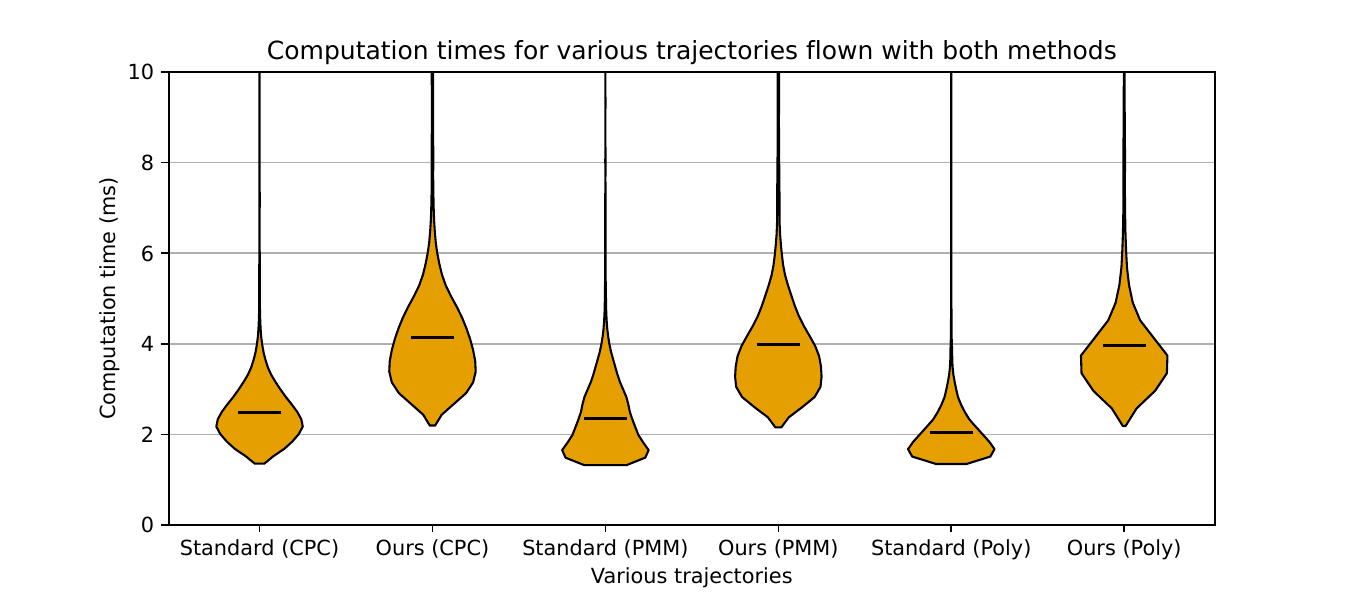}
        \vspace{-1.0em}
        \caption{Computation time statistics and comparison for various trajectories in the real-world (ARM architecture). }
		\label{fig:violin_plot_computation_time_realworld}
	\end{subfigure}
	\hfill
	\begin{subfigure}[b]{0.5\textwidth}
		\centering
		\includegraphics[width=\textwidth]{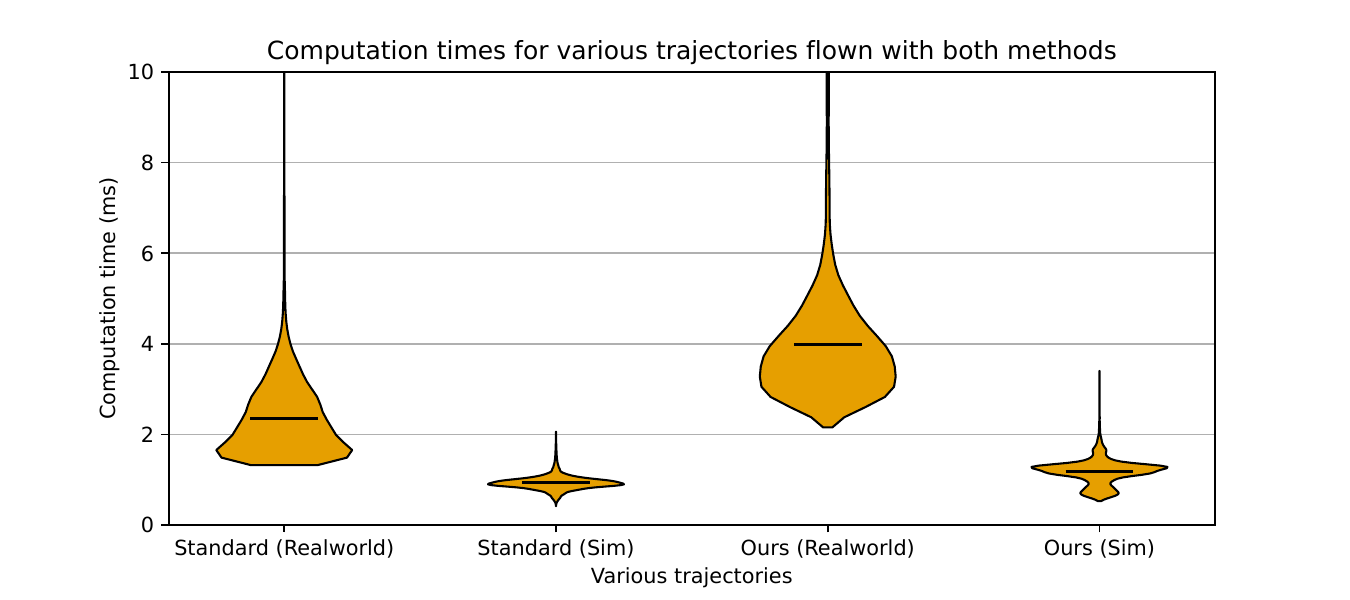}
        \vspace{-1.0em}
        \caption{Computation time comparison for flights in the real-world (ARM architecture) and simulations (AMD Ryzen 5850U).}
		\label{fig:violin_plot_computation_time_sim_vs_realworld}
	\end{subfigure}
	\caption{Computation time statistics for simulations and real-world. The black bar marks the mean for each case.}
    \vspace{-1.5em}
	\label{fig:violin_plot_computation_time}
\end{figure}

\begin{figure}[h!]
		\centering
		\includegraphics[width=0.5\textwidth]{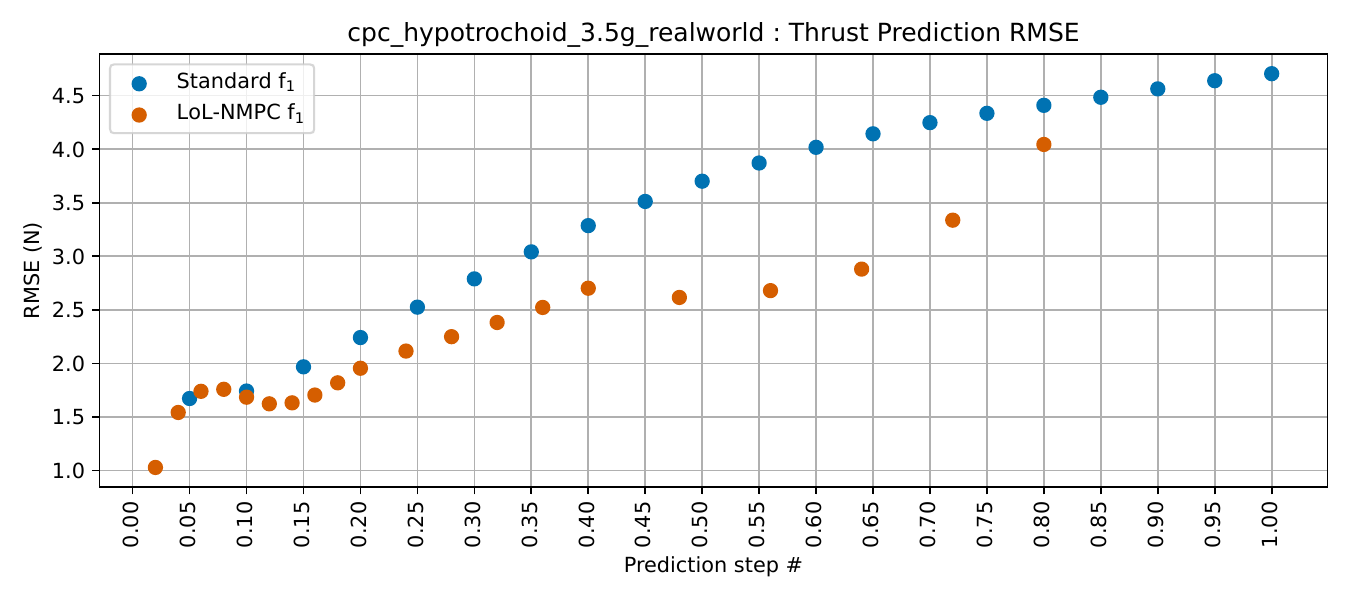}
        \vspace{-1.8em}
	\caption{Comparison of proposed and standard method for prediction of thrusts in real-world tests.}
    \vspace{-1.3em}
    \label{fig:real_world_prediction_comparison}
\end{figure}

\begin{figure}[h!]
	\centering
	\includegraphics[width=0.44\textwidth]{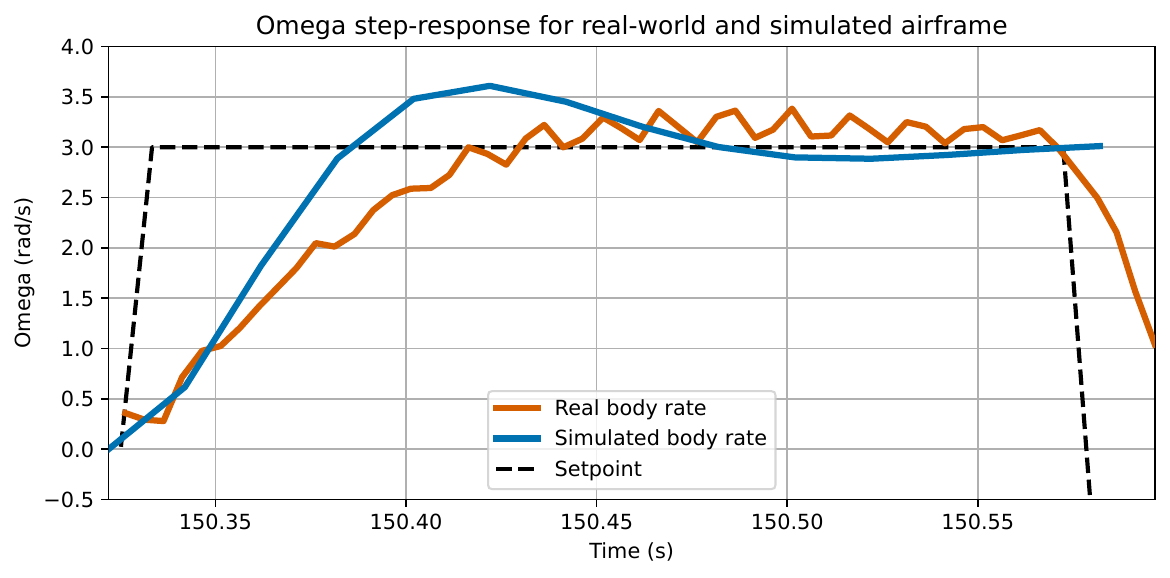}
    \vspace{-1.0em}
	\caption{Comparison of body rate response of the controller in simulation and in the real-world.}
    \vspace{-1.5em}
	\label{fig:omega_response_comparison}
\end{figure}



\subsection{Simulation Results}

\paragraph{Computation time} A primary requirement of high-speed agile manoeuvres is real-time decision making, and for this purpose, the high-level controller is expected to run at \SI{100}{\hertz}.
The predictive controllers have been limited in model fidelity due to this computation time limit of \SI{10}{\milli\second}, and through \reffig{fig:violin_plot_computation_time_sim_vs_realworld}, we show that our proposed model meets the requirement for computation limit both in a simulation on a laptop (with AMD Ryzen 5850U), but more importantly on an ARM-based onboard companion computer of the \ac{uav}.

\paragraph{Tracking Error} For simulation environment, we present the statistical analysis across 180 different flight tests conducted for each controller (10 flights per test case) through \reftable{table:sim_results_trifecta}.
The \ac{rmse} error was calculated over each trajectory flown and the table \reftable{table:sim_results_trifecta} shows the mean and standard deviation of the \ac{rmse} error over 10 runs performed per trajectory per controller.
The results clearly show the low variance and high stability in the performance of both controllers across all trajectories.
Additionally, it can be seen that our proposed controller outperforms the standard controller by up to \SI{29.16}{\percent} in tracking error, and offers an average gain of \SI{22.31}{\percent} across all trajectories.

\subsection{Real-world results}
For the presentation of real-world \ac{rmse} results, we first tested the real-world variability of the results by conducting several flights with the slanted figure-8 trajectory at \SI{3.5}{\g} using both the CPC and PMM methods.
The results showed a standard deviation of $\SI{0.1}{\meter}$ for CPC and $\SI{0.011}{\meter}$ for PMM, and therefore, we can conclude that the presented results are repeatable and show the true performance gained over the standard methods.

\paragraph{Tracking Error} Experimental results from the real-world flights are presented in \reftable{table:real_world_results_trifecta} and show the \ac{rmse} tracking error for each trajectory in its own row, and for each controller in its column.
The results indicate a clear performance improvement over the standard methodology with a maximum of \SI{38.6}{\percent} gain in performance, and an average gain of \SI{21.97}{\percent} across all trajectories.
The average gain in performance matches the predicted gain from simulation experiments, thereby proving the reliability of the proposed testing methodology in both simulated and real-world environments.

\paragraph{Prediction Error} Given the results from the earlier section, we present a detailed look at how the performance gain can be quantified and studied further.
For this purpose, we analyse the prediction performance of the controller to quantify the claimed higher fidelity offered by our proposed model inside the \textit{LoL-NMPC}.
For this analysis, the real-world data was interpolated to determine the timestamp closest to the predicted time step from every instance of the \ac{nmpc} flight, and \reffig{fig:real_world_prediction_comparison} show the \ac{rmse} of predictions made for position and thrust, respectively.
\reffig{fig:real_world_prediction_comparison} shows the clear advantage in prediction performance for actuator force during trajectory tracking, which arises from the novel actuator modelling inside our formulation.

\begin{table}[t!]

	\caption{RMSE for trajectory tracking in simulations}
    \vspace{-0.8em}
	\label{table:sim_results_trifecta}
	{
		\renewcommand{\tabcolsep}{4.5pt}
		\renewcommand{\arraystretch}{1.0}
		\centering
        \resizebox{0.5\textwidth}{!}{%
		\begin{tabular}{c  c  r  r  r r}
			\hline
			type & trajectory & \multicolumn{1}{c}{acc.} & \multicolumn{1}{c}{Standard (m)} & \multicolumn{1}{c}{ours (m)} & \multicolumn{1}{c}{gain (\%)} \\
			\hline                                                                                                         \\[-0.9em]
            CPC  & fig 8      & $2.5$g                   & $0.456 \pm 0.013$                & $\bm{0.370 \pm 0.008}$       & $18.85$\\
			CPC  & fig 8      & $3.5$g                   & $0.662 \pm 0.050$                & $\bm{0.477 \pm 0.018}$       & $27.94$\\
			CPC  & sl. fig 8  & $2.5$g                   & $0.457 \pm 0.007$                & $\bm{0.373 \pm 0.011}$       & $18.38$\\
			CPC  & sl. fig 8  & $3.5$g                   & $0.650 \pm 0.010$                & $\bm{0.489 \pm 0.036}$       & $24.76$\\
			CPC  & hyp.       & $2.5$g                   & $0.441 \pm 0.014$                & $\bm{0.384 \pm 0.016}$       & $12.92$\\
			CPC  & hyp.       & $3.5$g                   & $0.577 \pm 0.036$                & $\bm{0.472 \pm 0.017}$       & $18.19$\\
			\hline                                                                                                           \\[-0.9em]
			PMM  & fig 8      & $2.5$g                   & $0.393 \pm 0.012$                & $\bm{0.321 \pm 0.009}$       & $18.32$\\
			PMM  & fig 8      & $3.5$g                   & $0.538 \pm 0.009$                & $\bm{0.422 \pm 0.014}$       & $21.56$\\
			PMM  & sl. fig 8  & $2.5$g                   & $0.405 \pm 0.010$                & $\bm{0.314 \pm 0.008}$       & $22.46$\\
			PMM  & sl. fig 8  & $3.5$g                   & $0.567 \pm 0.016$                & $\bm{0.437 \pm 0.016}$       & $22.92$\\
			PMM  & hyp.       & $2.5$g                   & $0.401 \pm 0.011$                & $\bm{0.322 \pm 0.011}$       & $19.70$\\
			PMM  & hyp.       & $3.5$g                   & $0.551 \pm 0.013$                & $\bm{0.433 \pm 0.016}$       & $21.41$\\
			\hline                                                                                                           \\[-0.9em]
			Poly & fig 8      & $2.5$g                   & $0.210 \pm 0.005$                & $\bm{0.153 \pm 0.004}$       & $27.14$\\
			Poly & fig 8      & $3.5$g                   & $0.300 \pm 0.011$                & $\bm{0.226 \pm 0.006}$       & $24.66$\\
			Poly & sl. fig 8  & $2.5$g                   & $0.216 \pm 0.008$                & $\bm{0.153 \pm 0.007}$       & $29.16$\\
			Poly & sl. fig 8  & $3.5$g                   & $0.299 \pm 0.006$                & $\bm{0.224 \pm 0.011}$       & $25.08$\\
			Poly & hyp.       & $2.5$g                   & $0.209 \pm 0.005$                & $\bm{0.150 \pm 0.006}$       & $28.22$\\
			Poly & hyp.       & $3.5$g                   & $0.271 \pm 0.006$                & $\bm{0.217 \pm 0.007}$       & $19.92$\\
			\hline
		\end{tabular}
        }
	}
    \vspace{-2.2em}
\end{table}

\begin{table}[t!]
	\caption{RMSE for trajectory tracking in real-world}
	\vspace{-0.8em}
    \label{table:real_world_results_trifecta}
	{\centering
		\renewcommand{\tabcolsep}{3.0pt}
		\renewcommand{\arraystretch}{1.0}
		\begin{tabular}{c  c  r  r  r r r}
			\hline
			type & trajectory & \multicolumn{1}{c}{acc.} & \multicolumn{1}{c}{Standard (m)} & \multicolumn{1}{c}{ours (m)} & \multicolumn{1}{c}{gain (\%)} & \multicolumn{1}{c}{max v (m/s)} \\
			\hline                                                                                                                                                                           \\[-0.9em]
			CPC  & fig 8      & $2.5$g                   & $0.647$                          & $\bm{0.517}$                 & $20.130$                      & $22.051$                        \\
            CPC  & fig 8      & $3.5$g                   & $0.931$                          & $\bm{0.740}$                 & $20.530$                      & $26.666$                        \\
			CPC  & sl. fig 8  & $2.5$g                   & $0.818$                          & $\bm{0.568}$                 & $30.577$                      & $22.173$                        \\
            CPC  & sl. fig 8  & $3.5$g                   & $1.261$                          & $\bm{1.046}$                 & $17.069$                      & $26.154$                        \\
			CPC  & hyp.       & $2.5$g                   & $0.501$                          & $\bm{0.464}$                 & $7.429$                       & $22.522$                        \\
            CPC  & hyp.       & $3.5$g                   & $\bm{0.783}$                     & $0.813$                      & $-3.767$                      & $27.381$                        \\
			\hline                                                                                                                                                                           \\[-0.9em]
			PMM  & fig 8      & $2.5$g                   & $0.456$                          & $\bm{0.385}$                 & $15.551$                      & $20.587$                        \\
            PMM  & fig 8      & $3.5$g                   & $0.797$                          & $\bm{0.575}$                 & $27.884$                      & $25.468$                        \\
			PMM  & sl. fig 8  & $2.5$g                   & $0.675$                          & $\bm{0.640}$                 & $5.251$                       & $20.766$                        \\
            PMM  & sl. fig 8  & $3.5$g                   & $1.018$                          & $\bm{0.726}$                 & $28.645$                      & $23.286$                        \\
			PMM  & hyp.       & $2.5$g                   & $0.480$                          & $\bm{0.331}$                 & $30.904$                      & $20.237$                        \\
            PMM  & hyp.       & $3.5$g                   & $0.846$                          & $\bm{0.566}$                 & $33.086$                      & $24.993$                        \\
			\hline                                                                                                                                                                           \\[-0.9em]
			Poly & fig 8      & $2.5$g                   & $0.510$                          & $\bm{0.481}$                 & $5.649$                       & $13.821$                        \\
            Poly & fig 8      & $3.5$g                   & $0.665$                          & $\bm{0.420}$                 & $36.871$                      & $17.276$                        \\
			Poly & sl. fig 8  & $2.5$g                   & $0.564$                          & $\bm{0.346}$                 & $38.609$                      & $14.120$                        \\
            Poly & sl. fig 8  & $3.5$g                   & $0.629$                          & $\bm{0.460}$                 & $26.783$                      & $17.692$                        \\
			Poly & hyp.       & $2.5$g                   & $0.389$                          & $\bm{0.278}$                 & $28.541$                      & $11.798$                        \\
            Poly & hyp.       & $3.5$g                   & $0.465$                          & $\bm{0.345}$                 & $25.718$                      & $14.685$                        \\
			\hline
		\end{tabular}
	}
    \vspace{-2.4em}
\end{table}

%

\section{Conclusion}

In this paper, we introduced a novel \ac{nmpc} formulation which integrates the usually neglected low-level controller dynamics in addition to drag and actuator dynamics.
This approach eliminates the need for a separate control re-allocation and mixing strategy for agile manoeuvres performed at throttle limits. 
Our proposed formulation allows the \ac{uav} to perform agile manoeuvres at very high speeds without compromising tracking performance at accelerations of up to \SI{3.5}{\g}. 
Through extensive validation and testing with over 350 flights in simulated and real-world environments, we demonstrate an average performance gain of $\approx$\SI{22}{\percent}, and a maximum performance gain of $\approx$\SI{38}{\percent} in tracking error, compared to standard methods in the real-world, all without external state estimators. 
Our method remains computationally feasible while offering a significant uplift in performance over the standard approach across all scenarios.



\bibliographystyle{IEEEtran}
\bibliography{ref.bib}
\end{document}